\documentclass[letterpaper]{article}
\usepackage{graphicx}
\usepackage{aaai}
\usepackage{times}
\usepackage{helvet}
\usepackage{courier}
\usepackage{url}
\usepackage{hyperref}
\usepackage[english]{babel}
\usepackage[nottoc]{tocbibind}

\begin{document}
\title{Solving Royal Game of Ur Using Reinforcement Learning}
\author{Sidharth Malhotra \hspace{5mm} Girik Malik\\
College of Computer and Information Sciences\\
Northeastern University\\
Boston, MA 02115\\
}
\maketitle
\begin{abstract}
\begin{quote}
Reinforcement Learning has recently surfaced as a very powerful tool to solve complex problems in the domain of board games, wherein an agent is generally required to learn complex strategies and moves based on its own experiences and rewards received. While RL has outperformed existing state-of-the-art methods used for playing simple video games and popular board games, it is yet to demonstrate its capability on ancient games. Here, we solve one such problem, where we train our agents using different methods namely Monte Carlo, Q-learning and Expected Sarsa to learn optimal policy to play the strategic Royal Game of Ur. The state space for our game is complex and large, but our agents show promising results at playing the game and learning important strategic moves. Although it is hard to conclude that when trained with limited resources which algorithm performs better overall, but Expected Sarsa shows promising results when it comes to fastest learning.
\end{quote}
\end{abstract}

\section{Introduction}
Reinforcement Learning (RL) \cite{rlbook} is a powerful optimisation method used for complex problems. In RL, an agent learns to perform a (set of) task(s) on the basis of how it has performed on the previous steps. The agent typically gets a reward for moving closer to the goal or the optimised value, and in some cases a punishment for deviating from its intended learning task. Reinforcement learning, in a lot of ways, is inspired by biological learning that generally happens in mammals; for e.g.: children learn a language by observing their environment and if they are able to mimic it well, they are rewarded with something in appreciation. Similar behaviour is also observed in animals, like dogs, which are given treats on successful completion of a task, say fetching a stick. Mammalian brain then tries and rewire itself so that it can perform certain actions which lead to successful completion of tasks and gives it some sort of short/long term rewards. 

A few algorithms in RL are also directly inspired by neuroscience and behavioural psychology. Temporal Difference learning (TD), for example, is one such example which has been rigorously studied in the mammalian brain, and the loss function in TD is known to perform in a way the brain spikes, in case of dopamine neurons, when given a reward \cite{b1} \cite{b2} \cite{b3} \cite{b4} \cite{b5}. Schultz et al. \cite{b6} reported that in case of a monkey rewarded with juice, the dopamine level shot up, when the reward was not expected, showing a difference in expected and actual rewards, as in the TD loss function. Overtime, this firing activity back propagated to the earliest reliable stimuli, and once the monkey was fully trained, the firing activity disappeared, and started resulting in a decline on when the expected reward was not produced. The field has not just benefitted unidirectionally, results from TD have also been used in the study of schizophrenia, and the effects of dopamine manipulation on learning \cite{b7}.

The community has always been interested in development of more sophisticated algorithms and applying them to real-life tasks. RL is not far-behind in this. With the practical viability of deep leaning, there have been significant progresses in training RL algorithms with deep learning and then applying them to solve problems with human-level accuracy. This has been lately demonstrated by the use of RL algorithms to train agents for playing atari games, where they have surpassed human accuracy on a wide array of games, without significant changes in the strategy \cite{atari}. Board games are also not left behind in this feat. Shortly after atari games, RL has been used to solve one of the most complex board games, and beat the world champion \cite{go}.

Traditional approaches on board games have failed on larger boards, given large state spaces, complex optimal value functions making it infeasible for the to learn using self-play, and underdeveloped algorithms, that have slower/skewed learning curves. The problem starts to become even more complex as soon as multi-agent strategies come into play. Cooperative v/s independent multi agent strategies have their own advantages and disadvantages \cite{b8}. Cooperative agents have shown improvements in scores at the cost of speed. Independent agents, on the other hand, have shown the reverse to be true. While the idea of having both agents have access to the other’s state and action space is an acceptable workaround, such a setting is not possible to have in a few cases, like card games and certain kind of board games.

In our paper we demonstrate the use of popular RL algorithms in playing the Royal Game of Ur. The use of RL algorithms to solve popular board games is not new \cite{go} \cite{b17} \cite{b18} \cite{b19} \cite{b20}, but the use of RL for solving Ur has not been attempted yet. The Game of Ur is known to be a fairly complex, two-player board game, with variable number of states and pawns. A complete background on Ur is given in section \ref{sec:gour}. We have compared the performance of on-policy Monte Carlo, Q-learning and Expected Sarsa on playing The Game of Ur, in an independent multi-agent setting. We compare these RL algorithms’ application to Ur, implemented in a simulator, similar to Open AI’s Gym \cite{b9}. For the implementation, we create our own simulator from scratch (see section \ref{subsec:goursim}, with similar functions as implemented in \cite{b9}. Our goal is to test the performance of these algorithms on the simulator, and for the agents to be able to achieve human level strategies. The algorithm is not provided any game-specific information or hand-designed features, and are not privy to the internal state of the simulator. Through the simulator, the algorithms are just given the state space and the possible actions to take, as well as the information that they already have from their previous actions. 

\section{Background}
\begin{figure}[]
    \centering
    \includegraphics[width=0.5\textwidth]{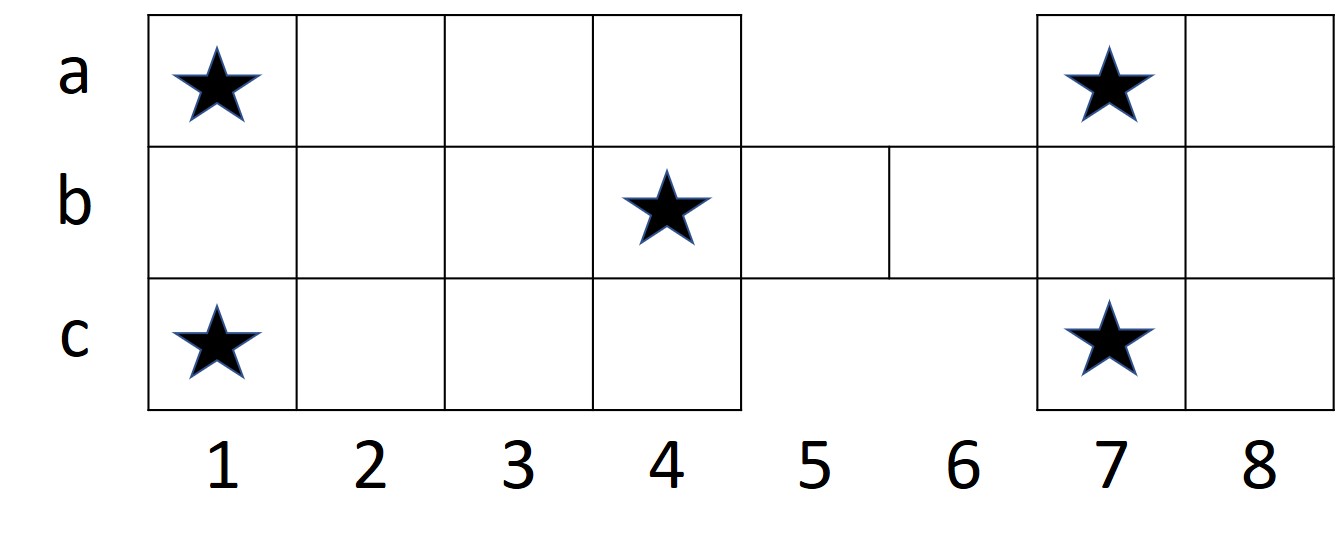}
    \caption{Board for The Royal Game of Ur, showing coordinates and safe positions.}
    \label{fig:board}
\end{figure}

\subsection{Game of Ur}
\label{sec:gour}
The Royal Game of Ur is a two player strategy racing board game, claimed to be first played in Mesopotamia in third century BC. However the exact origin of the game is still a matter of debate among historians. The game was long forgotten, except being rediscovered among the Jewish population in Indian city of Kochi, who continued playing a variant of it until 1950s. The Game of Ur is also believed to inspire/transformed into the early form of Backgammon \cite{b10}. 

The board consists of 20 squares, a large block of 3x4 squares, a small block of 3x2 squares, and a bridge of 1x2 blocks connecting the two. Each player starts with seven pieces (pawns) in hand and an empty board. There are four pyramidal dice with two marked and two unmarked corners. The number of marked corners that show up decide the number of positions to be moved by the active player. On a score of four, player can roll over the dice again. A pawn must bear off by exact throw. On landing on opponent’s pawn, the piece is removed from the board and must begin its journey again. A piece sitting on a special star-marked square is safe; the opponent can neither land on it, nor remove the other opponent’s piece from that position. The first four squares in the beginning of the pawn’s journey are safe by default, since an opponent cannot land there. The first player whose all seven pieces are all borne off the board has won the game \cite{b11}.

\begin{figure}[]
    \centering
    \includegraphics[width=0.5\textwidth]{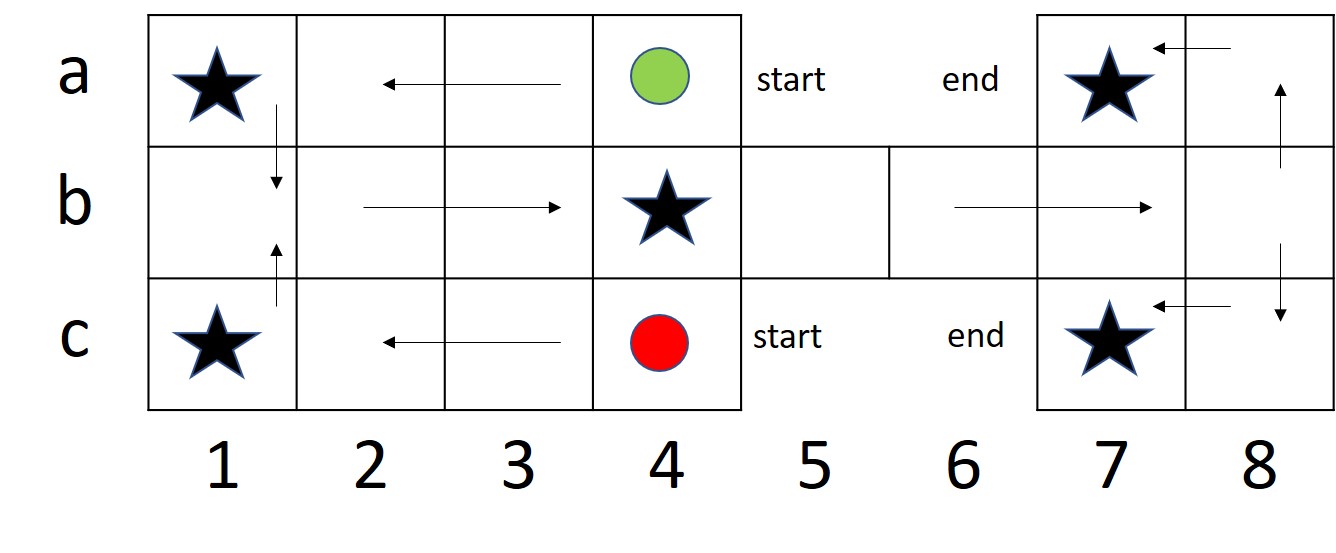}
    \caption{Board movements as happen during a normal gameplay. The circles in green and red represent different players and arrows represent the movement of respective pieces.}
    \label{fig:board_movements}
\end{figure}

\subsubsection{Gameplay}
The general move for the game is to move forward by the number of positions as shown up on the dice. The two players start on their respective start positions, on the opposite sides of the board, generally marked off as start. The rule for starting vary depending on the version and the place of game being played. In some places any number that shows up on the dice could be allowed to start the game, while in other places a particular number might be required to start-off with a particular piece. In yet another variant, the particular number on the dice might be used to just bring the piece on the board, and then another dice roll decides the number of pieces to be moved, whereas in some variants the second dice roll might not be allowed. The general movement of pieces is divided into war zones and safe zones. In case of a safe zone, the player has no risk of the piece being eliminated by the opponent, since the opponent cannot enter those positions (marked as rows a and c in the figure \ref{fig:board}), while in case of a war zone (marked as row b in the board, figure \ref{fig:board}), the player’s piece might get eliminated by the opponent, depending on the position of the piece and the number that shows up on the dice during opponent’s turn to roll the dice. The positions in the broad marked by a star are called safe positions, where the opponent cannot eliminate the other player’s piece. If a player’s piece is sitting on the safe position and the opponent’s piece lands on it, then the opponent has to move to the next position. There can be only a single piece situation on a position at any given time. In case if the colliding piece is that of the opponent, the piece gets eliminated, while if the piece is that of the current player, the move is not allowed. 

\subsection{Algorithms}
We also use the derivatives of Temporal Difference (TD) Algorithms, therefore it is important to give a background on TD algorithms \cite{b14}. 

\subsection{The Temporal Difference (TD) Algorithm}
The problem solved by reinforcement learning, which defines an error signal $\delta_t$ for state $V(s_t)$:
$$\delta_t = r(s')+\gamma V(s') - V(s)$$

The error signal vanishes if $V(s_t)$ reaches the target $r(s')+\gamma V(s')$. The algorithm waits for the next state $s'$ and looks if for that state the reward or the value function is known. If so, it changes $V(s)$ in that direction. Parameter $\gamma$ (typically 0.9) is called the discount factor. It takes into account that a state further away in future. say $V(s') = 1$, might be a possible successor of state $s$ in the course of the game, but it is not guaranteed that the game path for the entire game leads from $s$ to $s'$. Thus, the game states more distant in the future are \textit{discounted} in their effect on $V(s)$.
This clarifies the name TD, since for most states the reward is zero, and the error signal in most cases is the temporal difference in the value function. 

\subsection{Q-learning}
Q-learning \cite{b13} is a popular learning algorithm that can be applied to most sequential tasks to learn the state-action value. It is an off-policy learning algorithm, as the policy being learnt is not executed during training, instead a behaviour policy is used which is exploratory in nature. But the target action is the one predicted by the policy using the following update rule iteratively at each time step $t$:
$$Q(s,a) \leftarrow Q(s,a) + \alpha (r+\gamma Q(s',a') - Q(s,a))$$
where $a$ is the action that the behaviour policy predicts in the state $s$, whereas $A$ is the action that our current learnt policy predicts. The above update rule can be derived from the Bellman optimality equation for MDPs. 

\subsection{Expected Sarsa}
Expected Sarsa improves the on-policy nature of Sarsa. Since the update rule of Sarsa is dependent on the next action $a'$ it cannot converge unless the learning rate is reduced $(a \rightarrow 0)$ or exploration is annealed $(\epsilon \rightarrow 0)$, since $a'$ always has some degree of randomness. Expected Sarsa changes this with an update rule that takes the expected action-value instead of the action-value of $s', a'$: 
$$Q_{t+1}(s,a) = Q_t(s,a)+\alpha (r+\gamma \sum_{a'} \pi (a' | s') Q_t(s', a') - Q_t(s,a))$$
Since the update rule is now independent of the next action taken, and rather dependent on the expected action-value, Expected Sarsa can indeed converge \cite{b16}. In case of a greedy policy $\pi$, Expected Sarsa is the same as Q-learning. 

\section{Experimental Setup}
\subsection{The Game of Ur Simulator}
\label{subsec:goursim}
For the purpose of this paper, we implemented our own simulator for the game of Ur, which serves as a testbed for our algorithms and their comparisons. The simulator is open source and available at \url{https://github.com/sidharth0094/game-of-ur-reinforcement-learning} 

The theoretical details of the simulator are given below. 

\subsection{Game representation as MDP}
\subsection{State representation:}
Each square of the board is given a coordinate, for e.g.: (a,1), (a, 2), etc. (as shown in Figure \ref{fig:board}). Each state is represented by a tuple $(((num\_p_1),(p_1\cdots n)), ((num\_p_2),(p_2\cdots n)), dice): position\_piece\_to\_move$, where $num\_p_1$ represents the number of pieces of player $p_1$ that are yet to start, $p_1\cdots n$ represents the coordinates of pieces of player $p_1$ that are on board, here $n$ represents the total number of pieces of a player, dice represents the number that shows up on the dice, and $position\_piece\_to\_move$ represents the position of the piece of the current player to be moved by the number of steps that shows up on the dice. Likewise for player $p_2$. An example  state for a game with $4$ pieces and player $1$’s turn would be like:  $((2, ((a, 3), (a, 4))), (3, ((c, 3))), 1): 4 $, here $2$ represents that player $1$ has 2 pieces that are yet to start, which is followed by two tuples of positions of pieces that are already on board, the following $3$ represents the number of pieces of player $2$ that are yet to start, followed by the position of piece on board, while $1$ is the number that shows up on the dice, and $4$ represents the position of piece of player $1$ to be moved, which in this case is the one currently at $(‘a’, 4)$.

\subsection{Rewards:}
\label{subsec:rewards}
The reward function for our MDP representation works in the following way. The agent is given a reward of $+10$ when it successfully replaces the opponent’s piece, since the opponent has to start with that piece again and the path clears for the current player. The safe positions in the safe zones do not have a reward associated with them, since the agent is encouraged to be in war zone, while landing on the single safe position inside war zone has a reward of $+20$ associated with it, as the piece here is safe, and it is a strategic position to capture opponent’s pieces that cross it; the agent might want to keep its piece here as long as possible. We also associate a reward of $-1$ with any step inside war zone, other than the one to the safe position described above, as the piece is vulnerable to attack by the opponent, and should move out of war zone as soon as possible. Winning has a reward of $+100$ associated with it, while loosing has a reward of $0$. All other positions on the board, namely in safe zone, have a reward of $0$ as well. 

Initially, we had set the reward to be $+1$ for winning, and $0$ for loosing, but we had to change to such a discretised function because of problems when the agent tries to learn. In the absence of such a function, the agent takes a very large number of episodes in order to learn efficiently, which results in more compute time. Also, the agent is unable to learn the strategic moves well enough, given the strategy is only applicable to a few parts of the game, while the rest are independent of how the agent plays. We found that this approach of breaking the reward function for multiple smaller sub-tasks helps the agent learn faster and saves compute time. 

\begin{figure}[]
    \centering
    \includegraphics[width=0.5\textwidth]{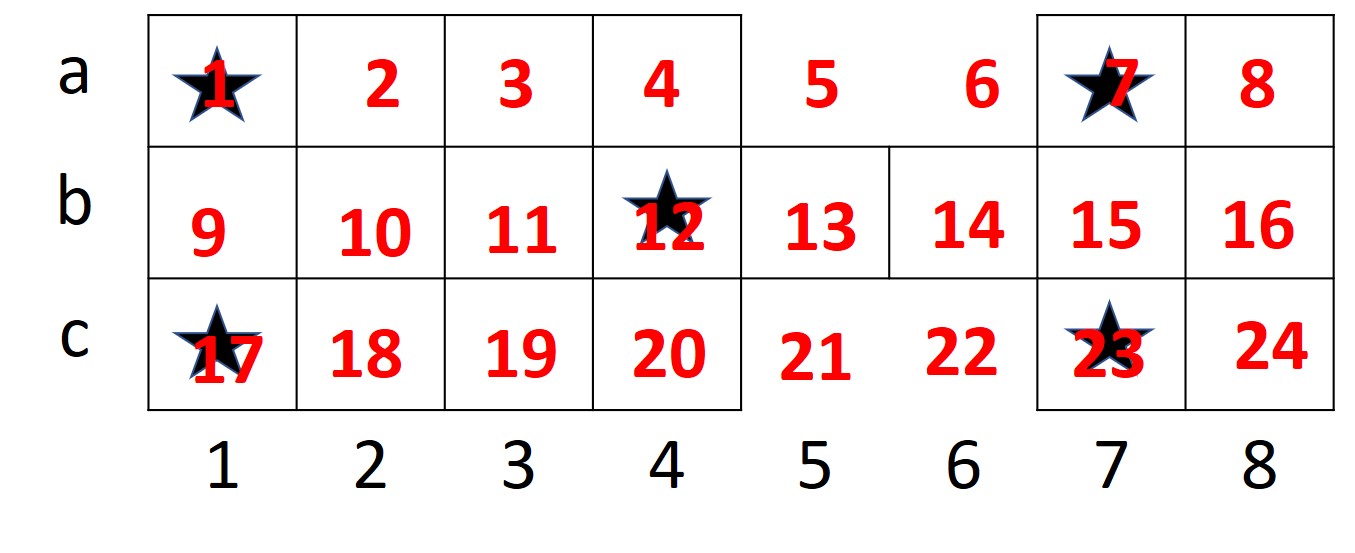}
    \caption{Encoding of board coordinates for faster training, used for representing board positions for deciding which piece to move forward}
    \label{fig:board_actions}
\end{figure}

\subsection{Actions:}
\label{subsec:actions}
All board positions are assigned an integer number, from 1 through 24, as shown in Figure \ref{fig:board_actions}, in order to decipher the piece that needs to move forward. This representation of using board positions to decide which piece to move forward is better than using individual piece IDs as it allows the agent to train faster by classifying the states with same coordinates, but different pieces, into a single state, as opposed to considering them as different states. The agent has the option to either go forward, or to make a null move, which happens in case there is no legal move. The situation of not having a legal move occurs in case all next positions are already occupied by the current player’s pieces, or the required number of steps to finish the game for a particular piece is less than the number that shows up on the dice.

\begin{figure}[]
    \centering
    \includegraphics[width=0.5\textwidth]{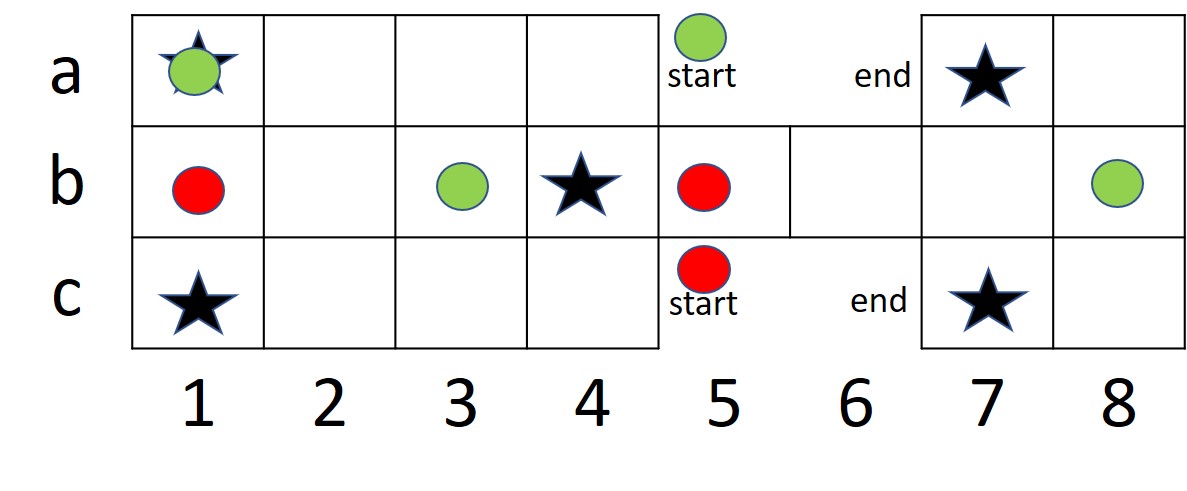}
    \caption{Strategic move when each player has $4$ pieces. Here, player represented by green piece has to decide between eliminating the red piece in $(‘b’, 1)$ or to move to a safe position in war zone. The decision making gets more complex with a full $7$ piece game, where green also has to avoid being eliminated by red. }
    \label{fig:strategy_move}
\end{figure}

\begin{figure}[]
    \centering
    \includegraphics[width=0.5\textwidth]{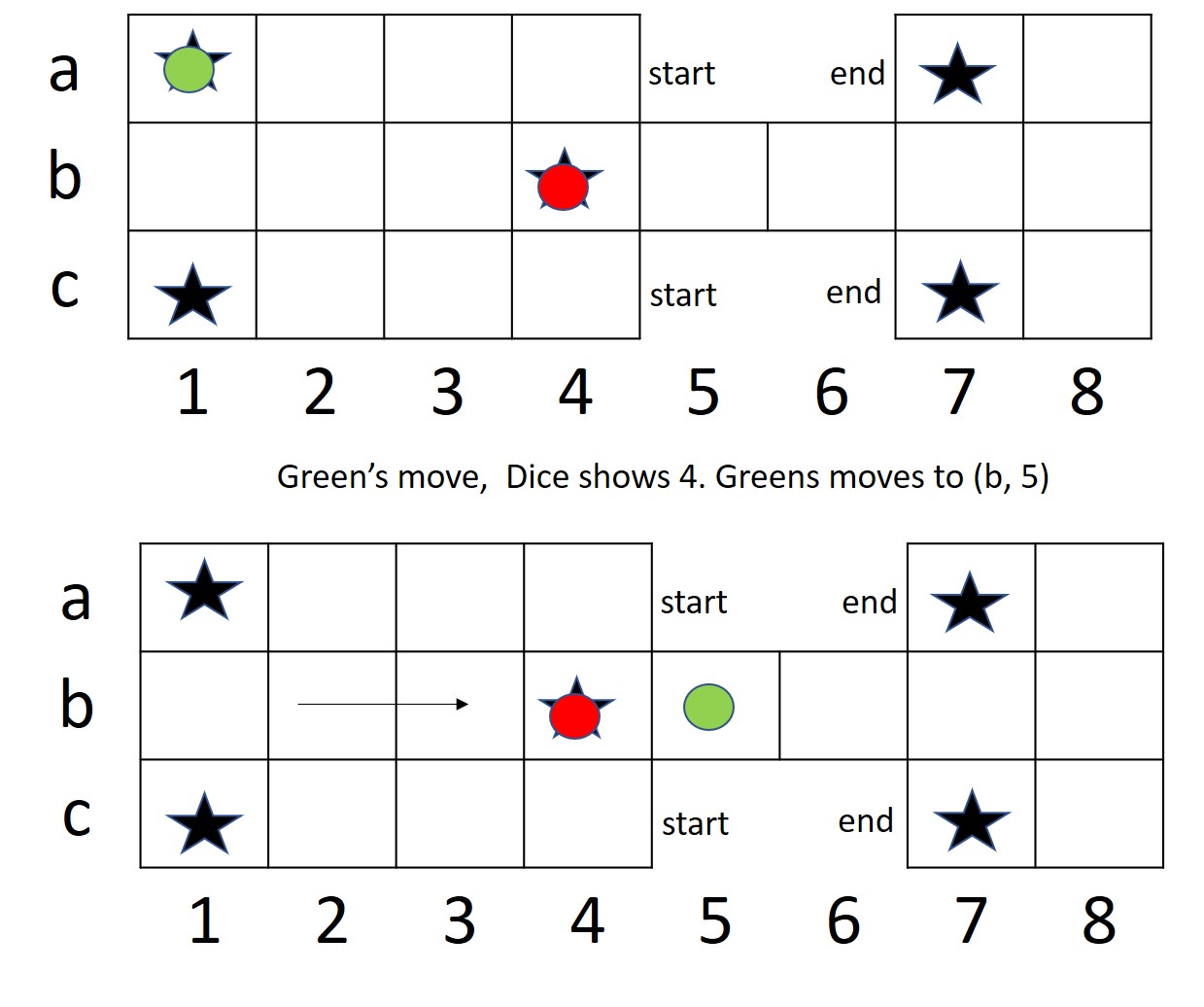}
    \caption{Safe move inside of a war zone. It is green’s turn to move, which with a dice roll of $4$ has to move to $(‘b’, 5)$. This move takes it closer to the goal state, and towards the end of war zone, as opposed to it starting with a new piece. Other pieces on board not shown in the figure}
    \label{fig:war_zone_safe_move}
\end{figure}

\begin{figure}[]
    \centering
    \includegraphics[width=0.5\textwidth]{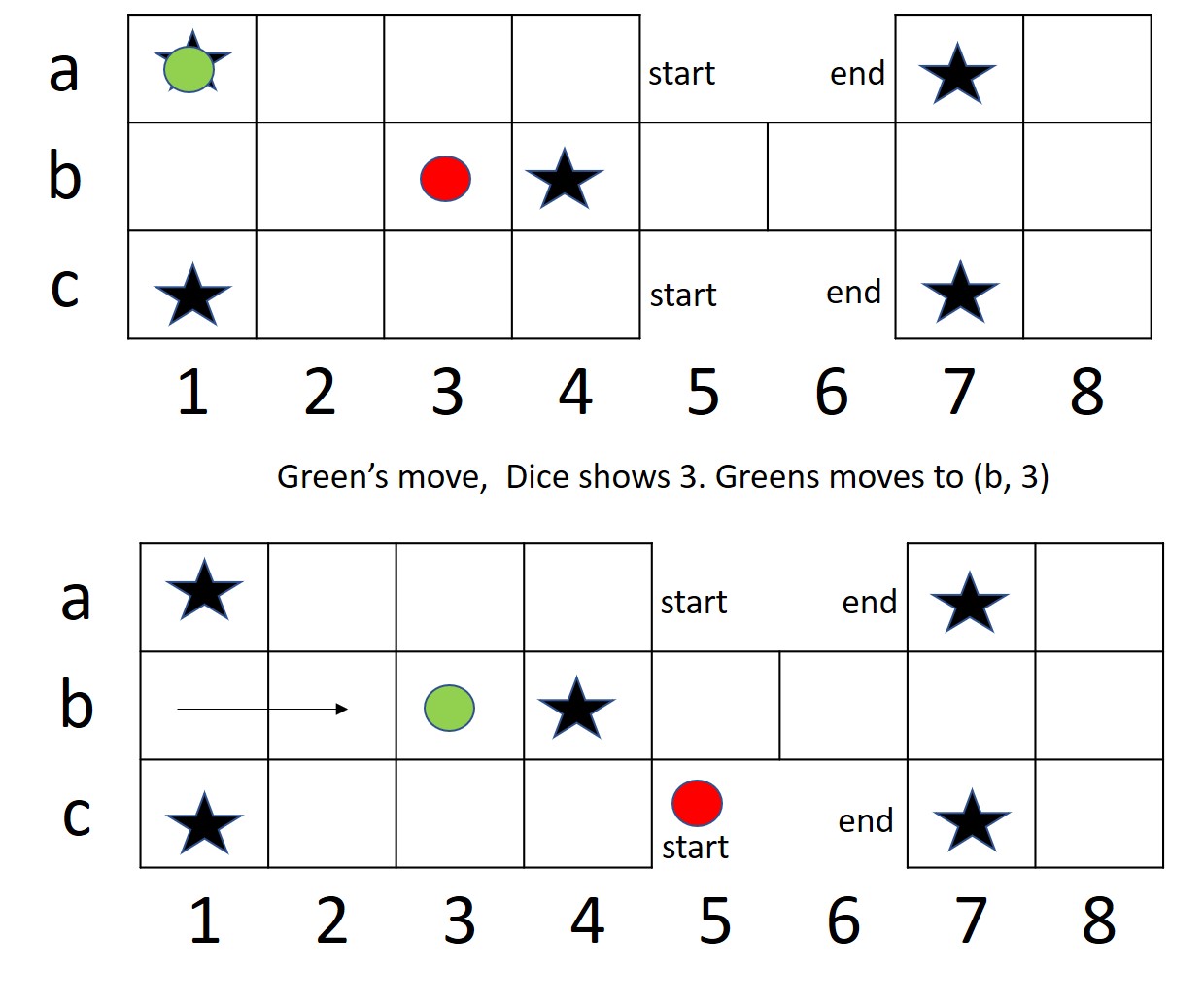}
    \caption{Unsafe move inside of a war zone. It is green’s turn to move, which with a dice roll of $3$ has to move to $(‘b’, 3)$. Here, red is vulnerable to be eliminated by green. Other pieces on board not shown in the figure}
    \label{fig:war_zone_unsafe_move}
\end{figure}

\section{Experiments}
We formulate the problem as a multiagent MDP, where both agents compete against each other to learn \cite{b12}. The reward function works as defined in (\ref{subsec:rewards}). For our experiments, we consider only two dice, since every board position is reachable by this configuration of dice. We have considered 4 pieces per player in order to train the agent quickly using limited computational resources at hand. Another reason to consider only 4 pieces per player is that this is the minimum number of pieces required for testing a popular strategic move, represented in figure \ref{fig:strategy_move}. Our action space consists of forward and null moves, as described in (\ref{subsec:actions}).

For our experiments, we used an \textit{$\epsilon$-greedy} approach, with an epsilon value of $0.1$, which signifies that the agent explores with a probability of $0.1$, while takes greedy actions with a probability of $0.9$. We trained our agent using Q-learning, Expected Sarsa, and on-policy Monte Carlo, since we thought it would be interesting to compare the performance of agents trained using episodic learning algorithms like on-policy Monte Carlo, against the popular TD learning based algorithms like tabular Q-learning and Expected Sarsa. Our agents are trained on $100$K episodes for each of the algorithm, and then tested on 100 gameplays against an agent following a stochastic policy with equiprobable actions. 

Since the state space of our environment is very large, therefore to keep a track of how the agent learns, we record the change in value function for state ((3, (('a', 3),)), (3, (('c', 3),)), 1), given that this state would occur frequently over the training period of all agents. We keep a track of time steps to win for each agent during training to observe if the agent is learning the strategic moves to finish early.

We also tested a special board position with 4 pieces (displayed in figure \ref{fig:strategy_move}), wherein we tried testing for what piece does agent decide to move based on the strategy it should learn. We show both safe and unsafe moves when the green piece enters the war zone in figures \ref{fig:war_zone_safe_move} and \ref{fig:war_zone_unsafe_move}. 

\section{Results}
\begin{table}[h]
\begin{tabular}{|l|l|l|}
\hline
\hline
\textbf{Algorithms}     & \textbf{Games won} & \textbf{Games lost}  \\
\hline
\hline
\textbf{Q-learning}     & 60                           & 40                      \\
\hline
\textbf{Monte Carlo}    & 55                           & 45                     \\
\hline
\textbf{Expected Sarsa} & 54                           & 46       \\
\hline
\end{tabular}
\caption{Table summarizing the results of agent performances trained on all three methods and then tested on $100$ gameplays.}
\label{tab:results}
\end{table}

We show the results of our testing for $100$ gameplays after training 3 separate agents using Monte Carlo, Q-learning and Expected Sarsa, summarized in Table \ref{tab:results}. For our algorithms, Q-learning wins $60$ out of $100$ games, while Monte Carlo and Expected Sarsa win $55$ and $54$ games respectively. This result is not completely random, and demonstrates an agent learning to play using popular strategic moves, as described below and shown in plots. 

\begin{figure}[]
    \centering
    \includegraphics[width=0.5\textwidth]{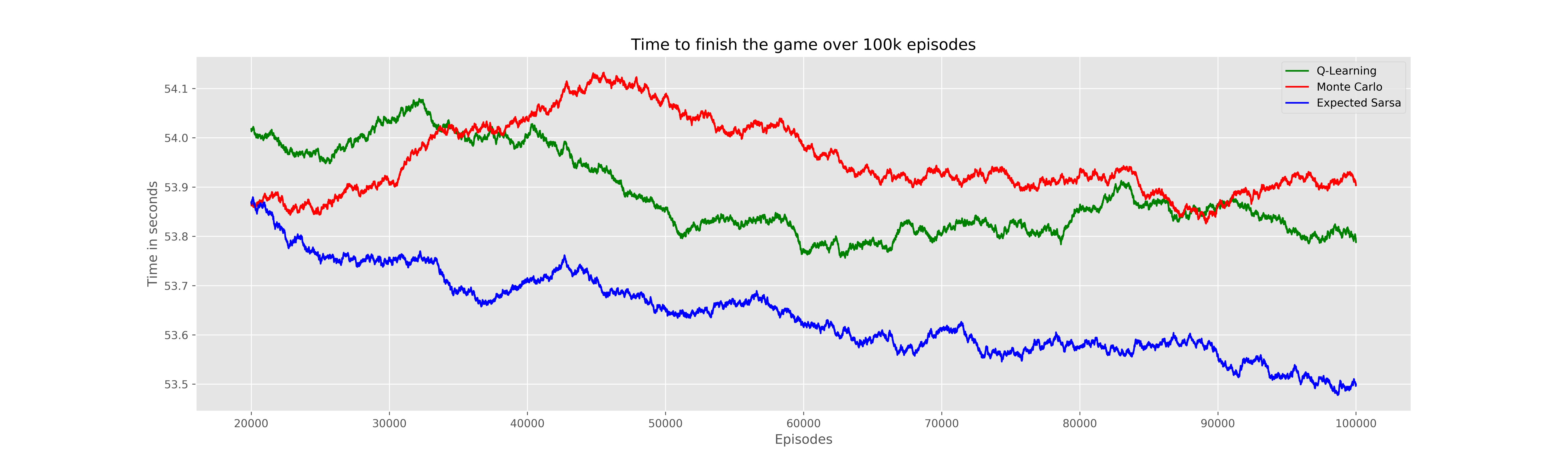}
    \caption{Time to finish for agents trained using the three methods. Here Q-learning, Monte Carlo and Expected Sarsa are represented by green, red and blue curves respectively. Note that Expected Sarsa demonstrates faster learning.}
    \label{fig:finish_times}
\end{figure}

We demonstrate the learning of our agents using the metric time to finish, as shown in figure \ref{fig:finish_times}. We observe that for all $3$ agents trained, our time to finish decreases over $100$K episodes. The sharpest decrease is shown by Expected Sarsa, while both Q-learning and Monte Carlo show similar competing curves. The curves do show fluctuations, but trend seems to move towards stabilization.

\begin{figure}[]
    \centering
    \includegraphics[width=0.5\textwidth]{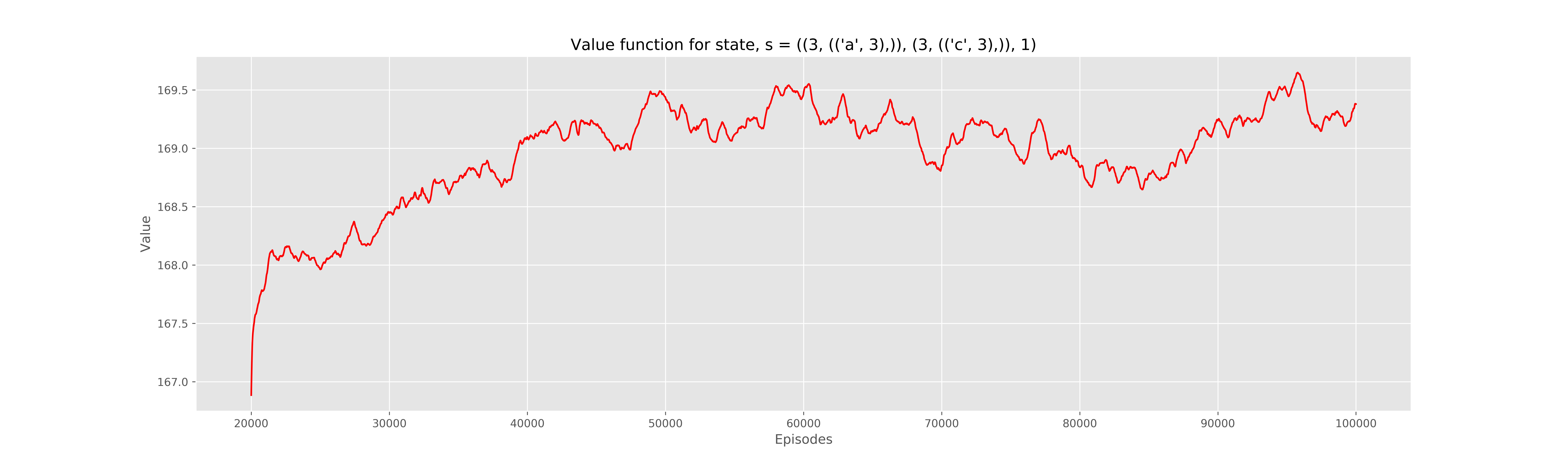}
    \caption{Value function for agent trained using Monte Carlo on state $((3, ((a, 3),)), (3, ((c, 3),)), 1)$. Note the sharp increase and fluctuating behaviour demonstrated.}
    \label{fig:value_function_mc}
\end{figure}

\begin{figure}[]
    \centering
    \includegraphics[width=0.5\textwidth]{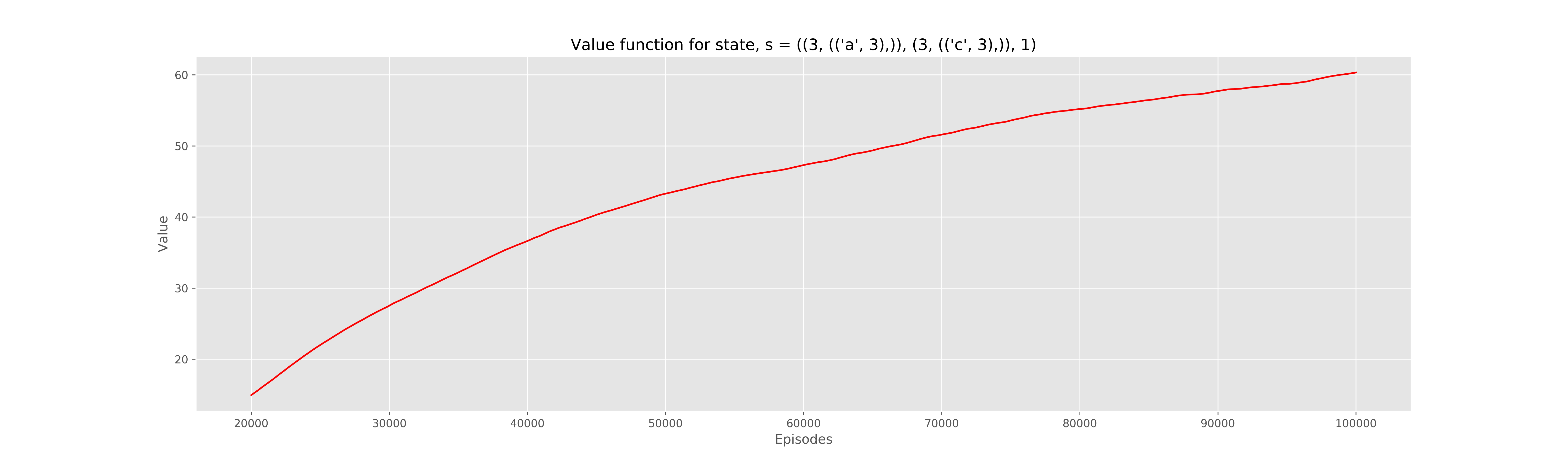}
    \caption{Value function for agent trained using Q-learning on state $((3, ((a, 3),)), (3, ((c, 3),)), 1)$.}
    \label{fig:value_function_q_learning}
\end{figure}

\begin{figure}[]
    \centering
    \includegraphics[width=0.5\textwidth]{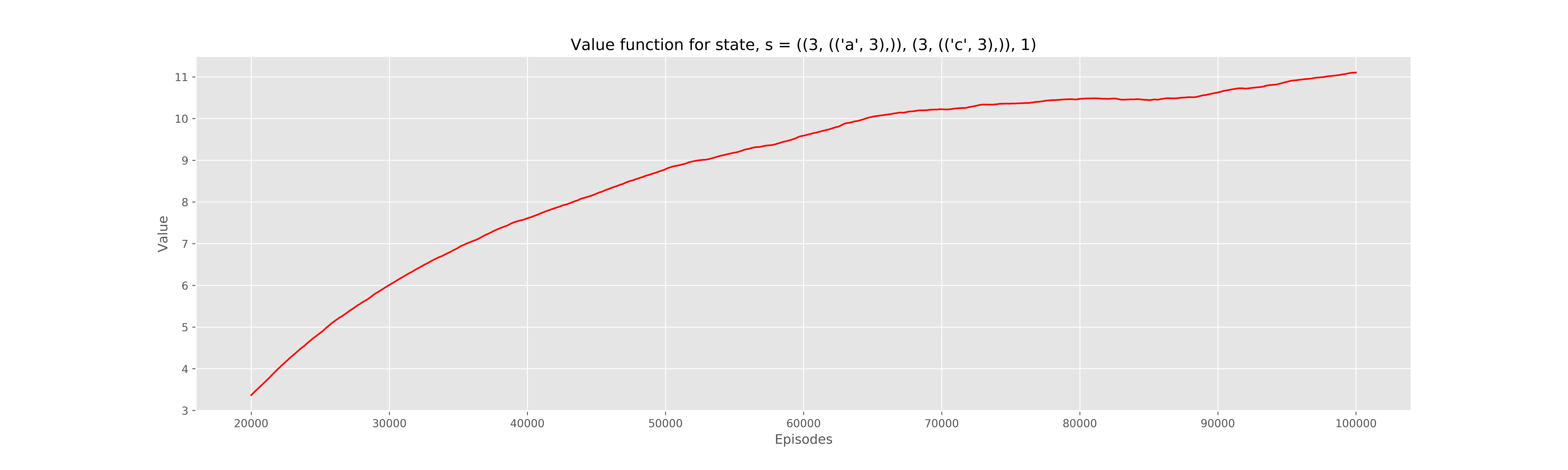}
    \caption{Value function for agent trained using Expected Sarsa on state $((3, ((a, 3),)), (3, ((c, 3),)), 1)$.}
    \label{fig:value_function_ex_sarsa}
\end{figure}

\begin{figure}[]
    \centering
    \includegraphics[width=0.5\textwidth]{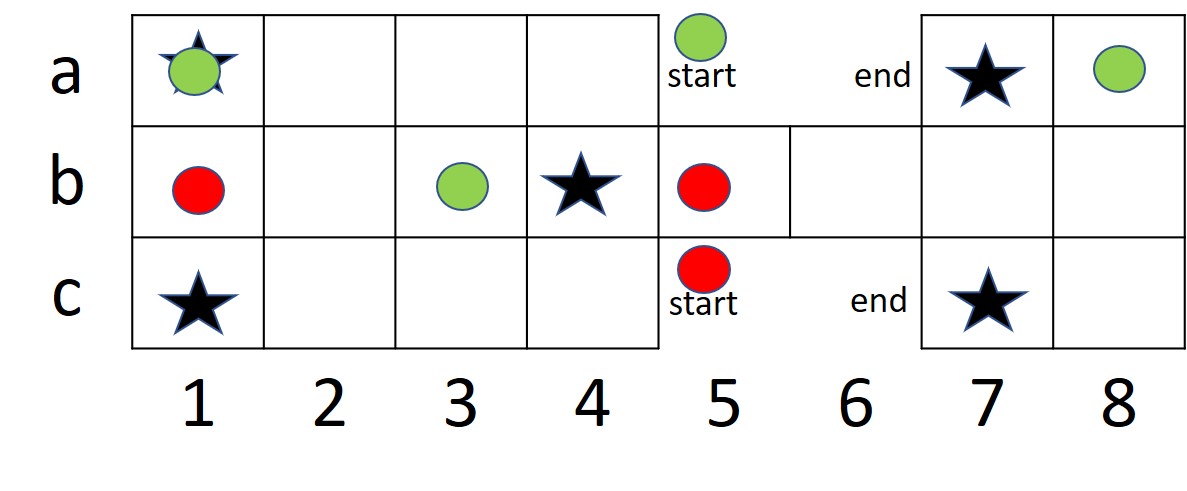}
    \caption{Strategic move learnt by the agent to save the piece from elimination, by taking it from $(‘b’, 8)$ to $(‘a’, 8)$, and saving it from opponent’s piece at $(‘b’, 5)$}
    \label{fig:strategy_move2}
\end{figure}

Our value functions for Monte Carlo, Q-learning and Expected Sarsa, for state $((3, ((a, 3),)), (3, ((c, 3),)), 1)$ are shown in Figures \ref{fig:value_function_mc}, \ref{fig:value_function_q_learning}, and \ref{fig:value_function_ex_sarsa} . The value function for the given state does seem to increase for all $3$ agents. In case of Monte Carlo, it shows a sharp increase, followed by a trend towards stabilization, while the plots for Q-learning and Expected Sarsa show a much smoother trend. One should not be misled that one agent is performing better over the other, it just shows that there is a difference in the way they learn. 

We also demonstrate the strategic move that our agents learn, as shown in figure \ref{fig:strategy_move2}. Our agents are able to learn this strategic move, in which the piece on coordinate $(‘b’, 8)$ is moved to $(‘a’, 8)$. This is an important move given the agent’s gameplay when at the intersection of war and safe zones. 

\section{Discussions}
The testing of our agents using different methods show promising results. The outcome was not always the smoothest, but for a lot of things we cannot conclude with certainty as to why an agent behaves in the given way. We can attribute this problem to the limitation of computational resources required for training our agents, with such complex and large state spaces. We believe that our agents could perform much better when trained with more episodes and better computational resources. 

\begin{figure}[]
    \centering
    \includegraphics[width=0.5\textwidth]{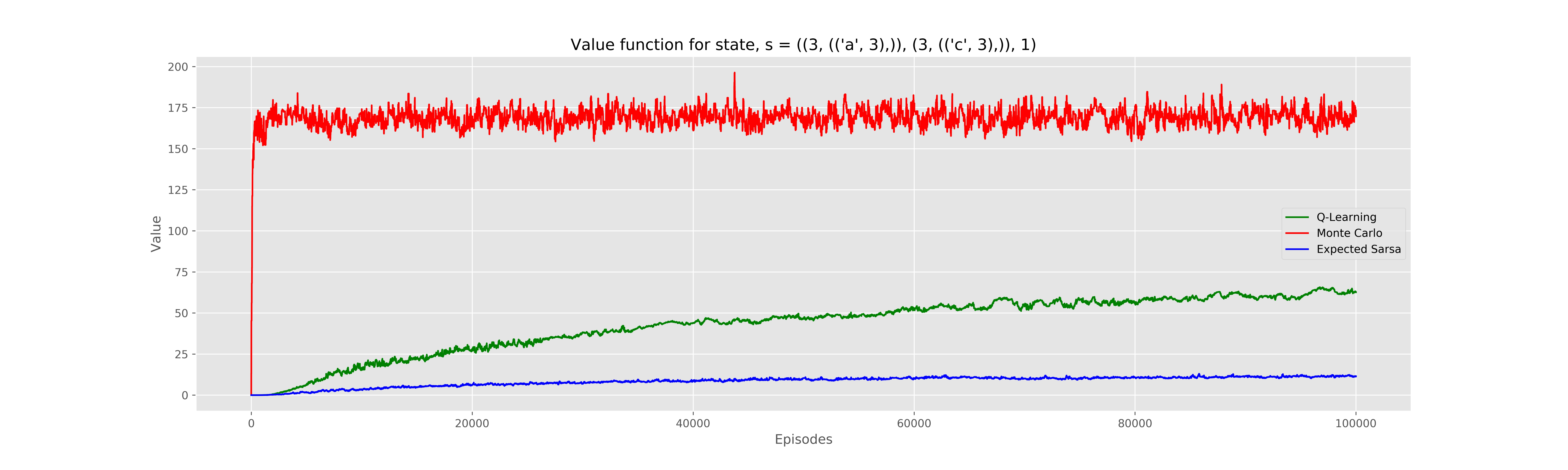}
    \caption{Combined value functions for agents trained using all three methods on state $((3, ((a, 3),)), (3, ((c, 3),)), 1)$. Q-learning, Monte Carlo and Expected Sarsa are shown by green, red and blue curves respectively.}
    \label{fig:q_learning_vs_mc_vs_ex_sarsa_values}
\end{figure}

We chose to show the value function for the state $((3, ((a, 3),)), (3, ((c, 3),)), 1)$, given that it is a prime state and occurs very frequently (comparison of all three methods together is shown in figure \ref{fig:q_learning_vs_mc_vs_ex_sarsa_values}). The disparity in smoothness, and the difference in values of the value functions for the given state, of our plots could be attributed to the fact that Monte Carlo takes full episode to learn, while TD methods do not. The step updates in case of TD methods are biased on the initial conditions of learning parameters. The bootstrapping process typically updates a function or lookup $Q(s,a)$ on a successor value $Q(s',a')$ using whatever the current estimates are in the latter. Clearly, at the very start of learning, these estimates contain no information from any real rewards or state transitions. However, if the agent is learning as it should, then this bias will reduce asymptotically over multiple iterations, but this bias is known to cause problems, specially for off-policy methods. Monte Carlo methods, on the other hand, do not suffer from this bias, as each update is made after the entire episode, using the true sample of $Q(s, a)$. But, Monte Carlo methods can suffer from high variance, which means more samples are required to achieve the same degree of learning compared to TD methods. A middle ground between these two problems could be achieved by using TD($\lambda$). 

Our agent learns to move the piece on $(‘b’, 8)$ inside of the war zone, to coordinate $(‘a’, 8)$ inside the safe zone. We believe that this is an important strategic move that it learns, since the piece at $(‘b’, 8)$ will reach the end position in two steps; while if the opponent’s piece at position $(‘b’, 5)$ eliminates it, then the piece at $(‘b’, 8)$ would have to restart. So agent by moving piece to $(‘a’, 8)$ didn’t just move it closer to winning state but also saved it from eliminating. 

\section{Conclusion}
In this report, we compared the performance of $3$ agents, trained using entirely different methods, namely Monte Carlo, Q-learning and Expected Sarsa, to play the ancient strategic board game, Royal Game of Ur. The state space for our game is complex and large, but our agents show promising results at playing the game and learning important strategic moves. Although it is hard to conclude that when trained with limited resources which algorithm performs better overall than the others, but Expected Sarsa showed promising results in case of fastest learning.

In future, we plan to run our agent on the given algorithms and their variants for more than $1$ million episodes, as is the case in community when testing on board games, as we speculate that this would allow our agent to experience more states, and therefore learn better policies.  We also plan to train our agent using sophisticated deep RL methods like DQN and Double DQN, to see if the agent shows significant differences in performance when trained on those.  


\bibliographystyle{apalike}
\bibliography{mybib}

\end{document}